\documentclass[times,twocolumn,final,authoryear]{elsarticle}

\usepackage{ycviu}
\usepackage{framed,multirow}

\usepackage{amssymb}
\usepackage{latexsym}
\usepackage{hyperref}
\hypersetup{
    colorlinks,
    linkcolor=newcolor,   
    citecolor=newcolor,  
    urlcolor=newcolor 
}

\usepackage{multirow}
\usepackage{bm}
\usepackage{colortbl}
\usepackage{array}
\usepackage{makecell}
\usepackage{tabularx}
\usepackage{dsfont}

\usepackage{url}
\usepackage{xcolor}
\definecolor{newcolor}{rgb}{.8,.349,.1}

\journal{Computer Vision and Image Understanding}

\begin{document}

\thispagestyle{empty}
                                                             
\clearpage
\thispagestyle{empty}
\ifpreprint
  \vspace*{-1pc}
\fi





\clearpage
\thispagestyle{empty}

\ifpreprint
  \vspace*{-1pc}
\else
\fi

\clearpage

\ifpreprint
  \setcounter{page}{1}
\else
  \setcounter{page}{1}
\fi

\begin{frontmatter}

    \title{A Turbo-Inference Strategy for Object Detection and Instance Segmentation}
    
    \author[1]{Zhen \snm{Zhao}}
    \author[2,3]{Gang \snm{Zhang}}
    \author[2,3,4]{Xiaolin \snm{Hu}}
    \author[1]{Liang \snm{Tang}\corref{cor1}}
    
    \cortext[cor1]{Corresponding author}
    \ead{happyliang@bjfu.edu.cn}

    \address[1]{School of Technology, Beijing Forestry University, Beijing 100091, China}
    \address[2]{Department of Computer Science and Technology, Tsinghua University, Beijing 100084, China}
    \address[3]{Beijing National Research Center for Information Science and Technology, Tsinghua University, Beijing 100084, China}
    \address[4]{Chinese Institute for Brain Research (CIBR), Beijing 100010, China}
    
    \received{1 May 2013}
    \finalform{10 May 2013}
    \accepted{13 May 2013}
    \availableonline{15 May 2013}
    \communicated{S. Sarkar}

    \begin{abstract}
    Object detection and instance segmentation tasks are closely related. Existing top-down instance segmentation methods usually follow a \textit{detect-then-segment} paradigm, where an initial detector is used to recognize and localize objects with bounding boxes, followed by the segmentation of an instance mask within each bounding box. In such methods, the detection accuracy directly influences the subsequent segmentation performance. However, previous research has seldom explored the impact of the instance segmentation task on object detection. In this \textit{paper}, we present a \textit{turbo-inference} strategy for the top-down methods that leverages the complementary information between detection and segmentation tasks iteratively. Specifically we design two modules: \textit{turbo-detection head} and \textit{turbo-segmentation head}, which facilitate communication between the tasks. The two modules form a closed loop that interlaces the detection and segmentation results without retraining the model. Comprehensive experiments on the COCO, iFLYTEK, and Cityscapes datasets demonstrate that our method substantially enhances both detection and segmentation accuracies with a certain increase in computational cost. The proposed method represents a tradeoff between prediction accuracy and inference speed.
    Codes are available at \url{https://github.com/zhaozhen2333/Turbo-Learning.git}.
    \end{abstract}
    

\end{frontmatter}


\begin{figure}[!t]
\centering
\includegraphics[width=0.48\textwidth]{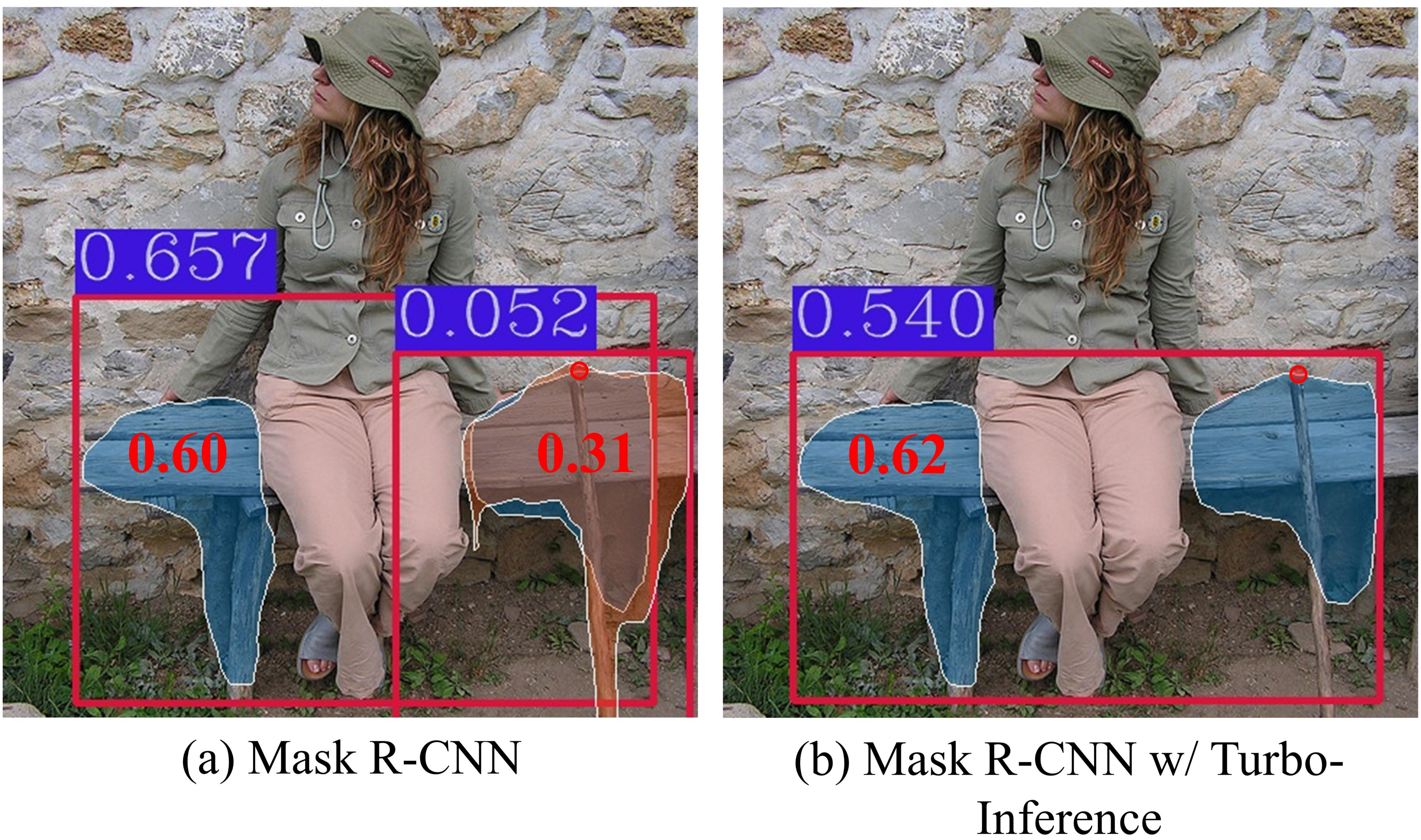}
\caption{
The predictions of the vanilla Mask R-CNN (a) and the Mask R-CNN with our turbo-inference strategy (b). 
Predicted masks are visually represented by randomly assigned colors, while detection boxes are highlighted in red. 
Classification scores are denoted as white numbers in a blue background. 
Red circles denote the extreme points of masks.
The mask quality scores, specifically, the Intersection over Union (IoU) values between predicted and human-annotated masks, are presented as red numbers.
}
\label{coco into}
\end{figure}

\section{Introduction}

Object detection and instance segmentation tasks aim to distinguish individual objects using bounding boxes and instance masks, respectively, and they are closely related. Existing instance segmentation methods can be categorized into top-down and bottom-up methods based on their reliance on object detection. Top-down methods ~\citep{MaskRcnn,maskscoring, BCnet,htc, lyu2022rtmdet, queryinstances,zhang2021refinemask,bolya2019yolact, tian2020conditional, vu2021scnet, zhang2020meinst} usually follow the \emph{detect-then-segment} paradigm, where an object detector is employed to generate bounding boxes for object localization and predict classification scores for object recognition, followed by a binary segmentation process (Figure \ref{coco into}(a)) within each detection box. Thanks to the advancements in object detectors, top-down methods have achieved state-of-the-art segmentation accuracy. In contrast, bottom-up methods~\citep{neven2019instancecut, solo, wang2020solov2, sparseinst, he2023fastinst, pei2022osformer} directly predict object masks from entire images, yielding high model efficiency. 
In this work, we aim to introduce new modules to further improve the performance of top-down methods.

In the top-down methods, the segmentation task is typically treated as a downstream task of object detection. The bounding boxes and classification scores obtained from the detection task play a crucial role in guiding the subsequent instance segmentation. However, these methods suffer from two problems. First, although pixel-level instance masks often contain more precise localization information compared with bounding boxes, these masks have not been effectively utilized to refine the coarse bounding boxes generated by the upstream detection task. For example, in Figure~\ref{coco into}(a), there is a significant distance between the extreme points (the highlighted red circles) of the instance masks and the bounding boxes. 
Second, classification scores are employed as a filtering metric in the evaluation process of both detection and segmentation tasks, however, the rich information contained in the segmentation task has not been leveraged to enhance the prediction of classification scores. 
For example, in Figure~\ref{coco into}(a), the redundant bounding box with a classification score of 0.052 has a very low mask quality score of 0.31. However, the redundant prediction is not corrected, leading to a decrease in accuracy.

We propose a \emph{turbo-inference strategy} to address the aforementioned issues. The turbo-inference strategy is characterized by its ability to integrate information from the downstream segmentation task, thereby refining the bounding boxes and classification scores produced by the upstream detection task. Specifically, the turbo-inference strategy introduces two modules: \emph{turbo-detection head} and \emph{turbo-segmentation head}. The turbo-detection head refines the initial bounding boxes and updates the classification scores to further eliminate low-quality redundant masks, utilizing the coarse masks predicted from the vanilla mask head. The turbo-segmentation head reuses the weights of the vanilla mask head and extracts Region of Interest (RoI) features based on the refined detection boxes output by the turbo-detection head, enabling more precise mask prediction. These two modules can be iteratively repeated to further enhance predictions. As a result, the detection and segmentation tasks form a closed-loop system, akin to a turbo-charger in an engine, where the outputs from one stage feed back into the subsequent stage, leading to a significant boost in overall performance.
In the examples shown in Figure~\ref{coco into}(b), the model with turbo-inference generates more precise target bounding boxes and eliminates redundant boxes. Then, in the next stage, it predicts higher-quality masks.
In addition, we found that this method can be combined with Soft NMS\citep{bodla2017soft} to further boost the performances of the tasks.

This idea was first used in a previous work\citep{turbo-learning} where the results of human-object interactions (HOI) recognition and pose estimation tasks are utilized to boost each other. One prominent difference between the two works is that in \citep{turbo-learning} the two tasks originally do not have dependence while in this work the two tasks are originally dependent (segmentation follows detection). We will show that the boosting strategy also works in this setting. Another prominent difference is that in \citep{turbo-learning} the boosting strategy is used in both training and inference, while in this work the strategy is only used in the inference stage, leaving the training cost unchanged.

We conducted extensive experiments on three datasets, including the widely used COCO~\citep{lin2014microsoft}, Cityscapes~\citep{cordts2016cityscapes} datasets, and the remote sensing dataset iFLYTEK~\citep{zhao2022winning}. The results showed that our method outperformed various baseline methods consistently, including the representative Mask R-CNN~\citep{MaskRcnn}, HTC~\citep{htc}, QueryInst~\citep{queryinstances} and RTMDet-ins\citep{lyu2022rtmdet}, in terms of both object detection and instance segmentation accuracies, demonstrating the effectiveness of our method.

\section{Related works}

\subsection{Object detection}

Object detection is a fundamental computer vision task that aims to recognize and detect objects with bounding boxes. Multi-stage detectors, represented by the R-CNN~\citep{R-CNN} family, have maintained state-of-the-art detection performance for a long period. Specifically, Fast R-CNN~\citep{fastrcnn}, Faster R-CNN~\citep{fasterrcnn}, and Cascade R-CNN~\citep{cascadercnn} first generate many candidate regions in the RPN (Region Proposal Network) stage, then use the detection networks to refine candidate boxes and distinguish instance categories in subsequent stages. In contrast, one-stage detectors directly predict locations and categories of objects without RPN, thus yielding high efficiency. The famous YOLO family~\citep{yolov1,redmon2018yolov3,ge2021yolox, bochkovskiy2020yolov4, li2022yolov6, xu2022pp, wang2023yolov7, ssda-yolo}, and FCOS~\citep{fcos, 3DF-FCOS} have exhibited competitive detection performance. In addition, the query-based set-prediction methods~\citep{vaswani2017attention, detr, Deformabledetr, zhang2022dinodetr, liu2022dab-detr, sun2021sparse}, such as DETR~\citep{detr}, Deforamble-DETR~\citep{Deformabledetr}, and Sparse R-CNN~\citep{sun2021sparse}, also achieve promising results and caught lots of attention recently.

\subsection{Instance segmentation}

Instance segmentation is an essential yet challenging vision task that requires assigning a pixel-level mask with an instance category for each object. Top-down instance segmentation methods achieve state-of-the-art segmentation performance. They typically follow the classical \emph{detect-then-segment} pipeline by designing additional segmentation networks to distinguish each pixel within detection boxes as foreground or background. For example, Mask R-CNN~\citep{MaskRcnn, faster_train_mask} introduces the FCN (Full Convolutional Network~\citep{fcn}) into the Faster R-CNN detector to segment objects in detection boxes. Mask Scoring R-CNN~\citep{maskscoring} and IoU Net~\citep{iounet} add an extra IoU head to re-score predicted masks and detection boxes, respectively. BMASK R-CNN~\citep{cheng2020boundary} and BCNet~\citep{BCnet} introduce additional contour supervision to pay more attention to boundary regions. 
In contrast, bottom-up instance segmentation methods~\citep{solo,wang2020solov2, sparseinst} are dedicated to improving real-time performance. 
Some of them are built upon semantic segmentation algorithms, which either project pixels into specific vector spaces or enable each pixel to learn an embedding to distinguish different instance clusters. There are also some attempts to directly handle instance segmentation. 
The representative Solo~\citep{solo, wang2020solov2}, and SparseInst~\citep{sparseinst} achieve promising segmentation accuracy while yielding high efficiency.

\subsection{Multi-stage refinement}
Multi-stage refinement has emerged as a popular paradigm for object detection and instance segmentation tasks. Cascade R-CNN~\citep{cascadercnn} introduces a sequence of detectors to achieve high-quality detection, where the output of each detector is fed into the next one for bounding box refinement. 
Sparse R-CNN~\citep{sun2021sparse} doubles the number of detectors by introducing learnable proposal boxes and proposal features within the Cascade R-CNN framework. HTC~\citep{htc} inherits the detection branch of Cascade R-CNN and designs an intertwined cascade mask branch to obtain more accurate instance masks. RefineMask~\citep{zhang2021refinemask} refines instance masks by incorporating fine-grained features iteratively during the multi-stage instance segmentation process. 
Different from the existing cascade structure within a single branch, we propose an interleaved cascade structure between detection and segmentation branches to refine both detection and segmentation results iteratively.

\begin{figure*}[!t]
\centering
\begin{minipage}[t]{0.97\textwidth}
\centering
\includegraphics[width=1\textwidth]{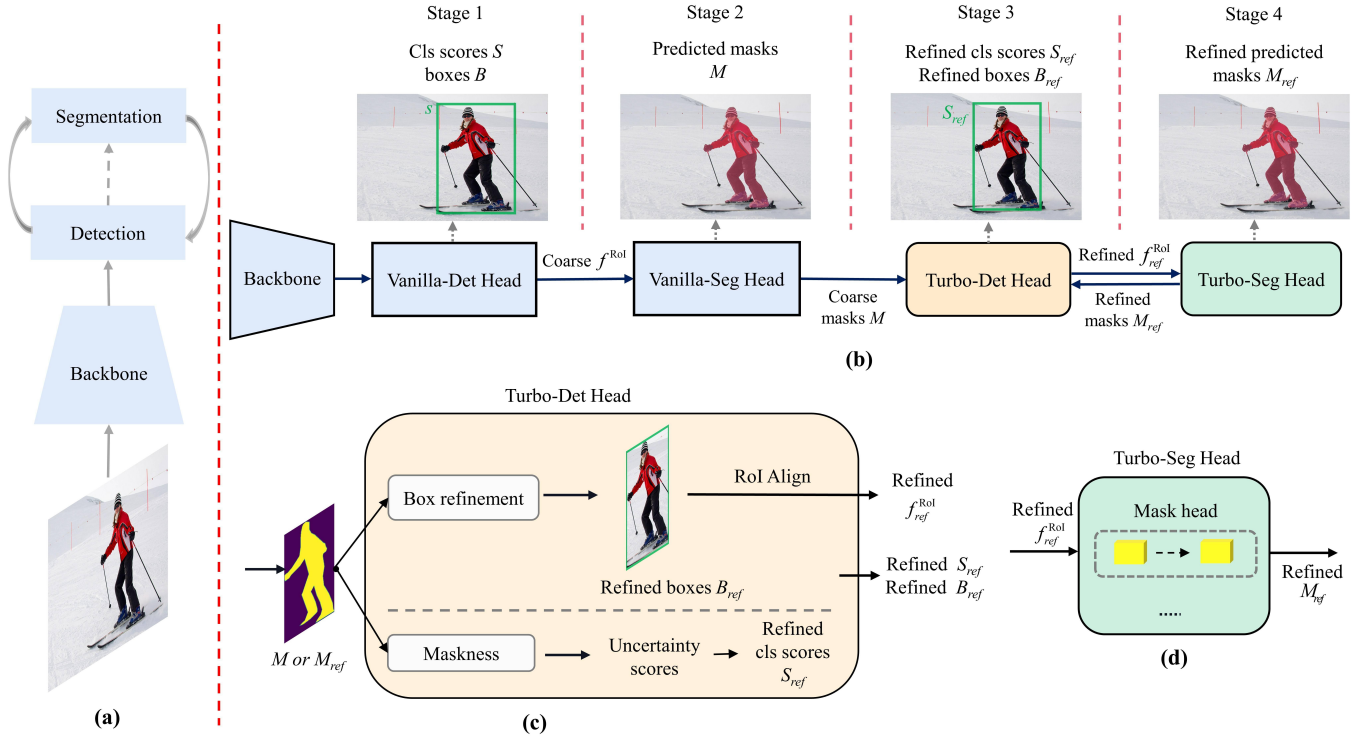}
\end{minipage}
\caption{Turbo-inference strategy. (a) Overview of the turbo-inference strategy. (b) Diagram of the turbo-inference strategy. Turbo-Det and Turbo-Seg are short for turbo-detection and turbo-segmentation. Vanilla \emph{detect-then-segment} pipeline constitutes stages 1 and 2, where the Vanilla-Seg head predicts coarse masks within detection boxes. In the Turbo-Det head in stage 3, the Box refinement module utilizes predicted coarse masks to generate tighter bounding boxes directly, while the maskness module leverages the uncertainty of instance masks to refine classification scores. After that, the refined bounding boxes are fed into the Turbo-Seg head in stage 4 to predict more precise instance masks. Stages 3 and 4 can be repeated multiple times for further refinement, leading to stages 5, 6 and so on. (c) The structure of the Turbo-Det head. (d) The structure of the Turbo-Seg head.
}
\label{TaskCascade}
\end{figure*}

\section{Method}

We begin with a well-trained model serving as our baseline. As a training-free method, turbo-inference strategy only modifies the inference stage of the baseline. 
Figure~\ref{TaskCascade}(a) and (b) show the pipeline of our proposed method. The turbo-inference strategy incorporates a turbo-detection head and a turbo-segmentation head to build a closed loop between the detection and segmentation tasks, enabling iterative refinement of both detection and segmentation results. Specifically, the coarse masks $M$ predicted by the vanilla-segmentation head are fed into the turbo-detection head to refine the detection boxes $B$ and classification scores $S$. Subsequently, the turbo-segmentation head is used to produce more precise instance masks $M_{ref}$ based on the refined bounding boxes $B_{ref}$. The enhanced instance masks can be fed into the turbo-detection head again.

\subsection{Turbo-detection head}

Figure \ref{TaskCascade}(c) depicts the turbo-detection head, which consists of two sub-modules: \emph{Box refinement} and \emph{Maskness}. The detection task comprises two sub-tasks, object localization and object classification, where the first sub-task outputs detection boxes $B$ and the second sub-task outputs classification scores $S$. The Box refinement module utilizes the localization information obtained from predicted masks $M$ to refine detection boxes, resulting in more accurate boxes $B_{ref}$. 
Meanwhile, the Maskness module leverages the spatial structure of predicted masks $M$ to refine classification scores $S$, obtaining more representative scores $S_{ref}$.


\subsubsection{Box refinement}
Figure \ref{coco example} shows that some predicted detection boxes may have excessive space and fail to enclose objects accurately, while predicted masks within boxes are more sensitive to the foreground and background pixels, containing more precise pixel-level localization information compared with detection boxes. However, masks predicted by the vanilla-segmentation head are typically encoded as floating-number maps with spatial size $m$$\times$$m$ to reduce training costs. Therefore, they do not contain explicit location information.

To address this issue, the Box refinement module employs bilinear interpolation operation~\citep{bilinear} to map predicted coarse masks back to the corresponding RoI region in the raw image, effectively decoding positional information. Subsequently, a threshold $S_{B}$ is employed to binarize the mapped masks to distinguish foreground and background pixels. Finally, the Box refinement module traverses all foreground pixels to calculate the maximum and minimum absolute values of coordinates, resulting in a refined box $[x_{min}, y_{min}, x_{max}, y_{max}]$. By incorporating detection boxes and predicted masks, our method enhances object localization accuracy, surpassing the approaches that use only detection boxes.

It is worth noting that using the commonly used threshold of 0.5 to distinguish between foreground and background pixels leads to a decrease in detection accuracy.
The potential reason for the decrease could be attributed to the presence of ambiguous foreground confidence regions (as mentioned in the subsequent section) within predicted masks. When the Box refinement module refines detection boxes using predicted masks, the commonly used foreground confidence threshold of 0.5 may discard these regions excessively. Consequently, the refined RoI regions may not encompass complete instances because the crucial information from these uncertain regions is overlooked.
In order to mitigate this problem, the generated bounding boxes are required to preserve moderately redundant spatial information and include the uncertain regions as much as possible.

\subsubsection{Maskness}

The Maskness module utilizes the uncertainty distribution of predicted masks to refine classification scores, thereby accurately filtering redundant predictions.
Previous research~\citep{layercascade} has shown that the majority of pixels in an image are relatively easy to distinguish, with foreground pixels having confidence scores close to 1 and background pixels close to 0. A tiny fraction of pixels play a crucial role in mask quality but exhibit ambiguous foreground/background confidence, with confidence scores close to 0.5. As mask quality increases, confidence scores within the foreground region are closer to 1. Therefore, the distribution of confidence scores in predicted masks can indicate the quality of masks.

We introduce \emph{uncertainty scores} to quantify the uncertainty distribution. The Maskness module initially applies a threshold $S_{M}$ to binarize predicted masks, distinguishing foreground and background regions. It subsequently calculates the cumulative confidence scores in foreground regions along with the total area of the foreground regions. The uncertainty score is defined as follows:
\begin{equation}
    U_\mathrm{mask} = \frac{\sum_i^N\mathds{1}(B_i > 0) M_i}{\sum_i^N B_i}
\end{equation}
\begin{equation}
    U_\mathrm{bbox} = \frac{(1 + U_\mathrm{mask})}{2}
\end{equation}
where $U_\mathrm{mask}$ and $U_\mathrm{bbox}$ are the uncertainty scores applied to predicted masks and detection boxes, respectively. $N$ denotes the number of all pixels. $M_{i}$ and $B_i$ denotes the confidence score and the binarized value in the $i$-th position.

The Maskness module assigns an uncertainty score to each mask and multiplies it with the corresponding classification score to generate a refined confidence score. We observed that uncertainty scores can provide valuable guidance for the detection box AP. However, they significantly reduce the $AP_{50}$ value, thereby weakening the detection box AP. Hence, we have reduced the magnitude of $U_{bbox}$ to achieve better detection gains.

\subsection {Turbo-segmentation head}

Figure \ref{TaskCascade}(d) illustrates the turbo-segmentation head, which shares network weights with the vanilla-segmentation head. Same as the vanilla-segmentation head, the turbo-segmentation head employs RoIAlign~\citep{MaskRcnn} to extract RoI features $f^\mathrm{RoI}_{ref}$ ($f^\mathrm{RoI}$) from the backbone features based on the location of detection boxes $B_{ref}$ ($B$). These RoI features are then propagated through multiple convolutions to predict instance masks $M_{ref}$ ($M$). Each value in these masks is a floating number within the range of 0-1, which denotes the confidence that a pixel belongs to the foreground. Finally, a threshold is used to binarize these masks.

\subsection{Turbo-inference strategy}

Better detection results lead to better segmentation results and better segmentation results inversely lead to more precise detection results. By introducing the turbo-detection head and the turbo-segmentation head, we extend the original two-stage pipeline into a multi-stage pipeline. The vanilla \emph{detect-then-segment} strategy constitutes stages 1 and 2 in our turbo-inference strategy. In stage 3, the turbo-detection head utilizes predicted masks from the vanilla-segmentation head to refine detection boxes and classification scores. In stage 4, the turbo-segmentation head leverages refined detection boxes to refine segmentation masks. The details are presented as follows:

1) In stage 1, the vanilla-detection head produces detection boxes $B$ and classification scores $S$. 

2) In stage 2, the vanilla-segmentation head predicts coarse masks $M$ based on the detection boxes $B$.

3) In stage 3, the turbo-detection head consists of two sub-modules: Box refinement and Maskness. The Box refinement module generates refined detection boxes $B_{ref}$ around \emph{predicted coarse masks} $M$. The Maskness module first calculates uncertainty scores $U_\mathrm{mask}$ and $U_\mathrm{bbox}$ by incorporating the uncertainty distribution of predicted masks. It then refines the classification scores $S$ by multiplying them with the uncertainty scores.

4) In stage 4, the turbo-segmentation head utilizes the refined detection boxes $B_{ref}$ to extract refined RoI features $f^\mathrm{RoI}_{ref}$, which are then used to predict refined predicted masks $M_{ref}$.


Stages 3 and 4 can be repeated multiple times, leading to stages 5, 6, and so on for further refinement.
Usually, introducing more inference stages would bring more improvement to both detection and segmentation results, but also increase the inference time. So we need a tradeoff between accuracy and computational cost. This can be determined empirically. 


\renewcommand{\arraystretch}{1.4} 

\begin{table*}[htbp]
    \caption{Main results on the COCO validation dataset. Turbo indicates turbo-inference. The inference time was measured on a single RTX 2080Ti GPU with batch size 2.}
    
    \begin{center}

    \begin{tabular}{c|c|p{0.6cm}p{0.6cm}p{0.6cm}p{0.6cm}p{0.6cm}p{0.6cm}|p{0.6cm}p{0.6cm}p{0.6cm}p{0.6cm}p{0.6cm}p{0.6cm}|c}

    \Xhline{2\arrayrulewidth}
    Method & Backbone & $\rm AP^b$ & $\rm AP_{50}^b$ & $\rm AP_{75}^b$ & $\rm AP_{S}^b$ & $\rm AP_{M}^b$ & $\rm AP_{L}^b$ & $\rm AP^m$ & $\rm AP_{50}^m$ & $\rm AP_{75}^m$ & $\rm AP_{S}^m$ & $\rm AP_{M}^m$ & $\rm AP_{L}^m$ & FPS\\
    
    \arrayrulecolor{black}\hline
    
    Mask R-CNN& R50-FPN&39.2  &\textbf{59.6}  &42.8 &22.9  &42.6  &51.2  &35.4   &56.4  &37.9 &\textbf{19.1}  &38.6  &48.4 & \textbf{15.7} \\
    w/ Turbo      & &\textbf{40.3}  &59.5  &\textbf{43.9} &\textbf{23.3}  & \textbf{43.9} & \textbf{53.5} &\textbf{36.7}  & \textbf{56.7} &\textbf{39.5} &17.2 &\textbf{39.3} &\textbf{53.7}&12.0 \\
    
    \arrayrulecolor{black}\hline
    
    Mask R-CNN& R101-FPN& 40.8 & \textbf{61.0} & 44.5 &23.0 &45.0 &54.1 &36.6&  57.9&  39.1 & \textbf{19.2} & 40.2 & 50.5& \textbf{13.5} \\
    
    w/ Turbo     & & \textbf{41.8} & 60.7 & \textbf{45.5} & \textbf{23.4} & \textbf{45.9} &\textbf{55.9} & \textbf{37.9}&  \textbf{58.0} &  \textbf{40.8} & 17.5&  \textbf{41.0} &  \textbf{55.3} & 9.8\\

    \arrayrulecolor{black}\hline
    
    Mask R-CNN& X101-FPN& 42.8 & \textbf{63.8}&  47.3& 25.4& 47.2& 55.9&38.4&  60.6&  41.3& \textbf{21.0} & 42.4& 52.2&\textbf{6.8} \\
    
    w/ Turbo & & \textbf{43.9} & 63.3 &  \textbf{48.0} &\textbf{25.8}&  \textbf{48.4}&  \textbf{57.5}& \textbf{39.7}&   \textbf{60.8}&   \textbf{42.8}& 19.4&  \textbf{43.1}&  \textbf{56.8}&5.5 \\
    
    \arrayrulecolor{black}\hline

    Mask R-CNN & CX-T-FPN& 46.2 & \textbf{68.1}&  50.8& 30.1& 49.5& 59.5&41.7 &  65.0&  44.9& \textbf{21.9} & 44.6& 59.9&\textbf{14.5} \\
    
    w/ Turbo & & \textbf{47.5} & 67.6 &  \textbf{51.7} &\textbf{30.6}&  \textbf{50.8}&  \textbf{61.9}& \textbf{42.8}&   \textbf{64.9}&   \textbf{46.3}& 22.5&  \textbf{45.8}&  \textbf{61.1}&10.8 \\
    
    \arrayrulecolor{black}\hline

    Mask R-CNN& Swin-T & 46.0 & \textbf{68.1}&  50.7 & 31.4 & 49.3 & 59.3 &41.7 &  65.2 &  44.4 & \textbf{23.5} & 44.9 & 59.7 &\textbf{9.8} \\
    
    w/ Turbo & & \textbf{47.5} & 67.8 &  \textbf{52.1} &\textbf{32.0}&  \textbf{51.1}&  \textbf{61.9}& \textbf{42.9}&   \textbf{65.2}&   \textbf{46.1}& 24.2&  \textbf{46.0}&  \textbf{61.1}&7.4 \\
    
    \arrayrulecolor{black}\hline

    Mask R-CNN& Swin-S & 48.2 & \textbf{69.8}&  52.8 & 32.1 & 51.8& 62.7 &43.2 &  67.1 &  46.1& \textbf{24.8} & 46.3 & 62.1 &\textbf{9.3} \\
    
    w/ Turbo & & \textbf{49.4} & 69.5 &  \textbf{53.8} &\textbf{32.9}&  \textbf{53.4}&  \textbf{64.6}& \textbf{44.5}&   \textbf{67.2}&   \textbf{47.7}& 25.4&  \textbf{47.7}&  \textbf{63.1}&6.3 \\
    
    \arrayrulecolor{black}\hline

    Mask R-CNN& Vit-B & 51.5 & \textbf{72.1}&  56.6 & 35.3 & 55.5& 66.4 &45.7 &  69.4 &  49.8& \textbf{26.9} & 49.0 & 64.1 &\textbf{3.9} \\
    
    w/ Turbo & & \textbf{52.0} & 71.8 &  \textbf{56.8} &\textbf{35.6}&  \textbf{56.0}&  \textbf{66.8}& \textbf{46.8}&   \textbf{69.4}&   \textbf{51.1}& 27.7&  \textbf{50.2}&  \textbf{64.9}&3.3 \\
    
    \arrayrulecolor{black}\hline

    Mask R-CNN& CX-v2-FPN & 52.9 & \textbf{72.7}&  58.7 & 35.9 & 57.3& 67.0 &46.4 &  69.9 &  50.7& \textbf{27.3} & 49.8 & 64.8 &\textbf{5.4} \\
    
    w/ Turbo &  & \textbf{53.4} & 72.4 &  \textbf{59.0} &\textbf{36.4}&  \textbf{57.9}&  \textbf{68.0}& \textbf{47.6}&   \textbf{69.8}&   \textbf{52.1}& 28.3&  \textbf{51.0}&  \textbf{65.7}&5.2 \\
    
    \arrayrulecolor{black}\hline
    
    HTC          & R50-FPN& 43.3 & \textbf{62.2}&  47.1 &24.3& 46.4& 57.7&38.3 & 59.3 & 41.4&\textbf{19.9} &41.0 &53.0& \textbf{5.5}\\
    
    w/ Turbo & & \textbf{43.7} & 62.1&  \textbf{47.5} &\textbf{24.4}& \textbf{46.8}& \textbf{58.5}&\textbf{39.2} &  \textbf{59.5} &  \textbf{42.4}&18.1& \textbf{41.5}& \textbf{57.6}&4.5\\
    
    \arrayrulecolor{black}\hline
    
    HTC          & R101-FPN& 44.8& \textbf{63.3} & 48.8& \textbf{25.7}& 48.5& 60.2&39.6&  61.0 & 42.8&\textbf{21.3}& 42.9& 55.0&\textbf{5.2} \\
    
    w/ Turbo & & \textbf{45.1} & 63.3&  \textbf{48.9}&25.6& \textbf{48.8}& \textbf{60.8}&\textbf{40.5}&  \textbf{61.1}&  \textbf{43.7}& 18.9& \textbf{43.5}& \textbf{59.4}&4.3\\
    \arrayrulecolor{black}\hline

    RTMDet-m& CSPX- & 48.8  & \textbf{66.8}  & 53.3& 29.9 & 54.0 & 65.1 &42.1 &  63.8  & 45.1 &19.3 & 46.4 & 63.1&\textbf{3.7} \\
    w/ Turbo & PAFPN& \textbf{49.3}  & 66.7 &  \textbf{53.5} &\textbf{29.9} & \textbf{54.5} & \textbf{66.1} &\textbf{42.4} &  \textbf{63.8} &  \textbf{45.5}& \textbf{19.5}& \textbf{46.7}& \textbf{63.3}&2.7\\
    \arrayrulecolor{black}\hline
    
    RTMDet-l& CSPX- & 51.1 & \textbf{69.0} & \textbf{55.8} & \textbf{32.9} & 56.1 & 67.7 &43.7 &  \textbf{66.0} & 47.0 &20.8 & 48.0 & 64.8 & \textbf{3.4} \\
    w/ Turbo &PAFPN & \textbf{51.4} & 68.9 & 55.6& 32.6 & \textbf{56.5} & \textbf{68.5} &\textbf{44.0}&  65.9 & \textbf{47.4} &\textbf{21.0} & \textbf{48.4} & \textbf{65.1} & 2.5 \\
    \arrayrulecolor{black}\hline
    
    RTMDet-x& CSPX- & 52.4  & \textbf{70.5}  & 57.2& \textbf{34.6} & 57.3 & 68.2 &44.6 &  67.4  & 47.8 &22.2 & 49.0 & 65.5&\textbf{2.8} \\
    w/ Turbo & PAFPN & \textbf{52.7}  & 70.4 &  \textbf{57.2} &34.4 & \textbf{57.5} & \textbf{69.1} &\textbf{44.9} &  \textbf{67.4} &  \textbf{48.4}& \textbf{22.6}& \textbf{49.3}& \textbf{65.8}& 2.2\\
    \arrayrulecolor{black}\hline
    
    QueryInst& R50-FPN& 42.0 & 60.5 & 45.6& 24.3 & 44.4 & 58.1&37.5&  58.7 & 40.5 &18.3 & 40.2 & 57.2&\textbf{7.5} \\
    w/ Turbo & & \textbf{42.8} & \textbf{60.7} &  \textbf{46.5}&\textbf{24.8} & \textbf{45.4}& \textbf{59.1} &\textbf{38.7}&  \textbf{59.3}& \textbf{41.9}& \textbf{19.0} & \textbf{41.1} & \textbf{58.1}&6.0\\
    \arrayrulecolor{black}\hline
    
    QueryInst& R101-FPN& 49.0  & 68.1  & 53.7& 31.5 & 51.9 & 64.3 &42.9 &  66.0  & 46.6 &23.5 & 46.3 & 62.0&\textbf{3.5} \\
    w/ Turbo & & \textbf{49.8}  & \textbf{68.2} &  \textbf{54.5} &\textbf{32.1} & \textbf{53.0} & \textbf{65.1} &\textbf{44.1} &  \textbf{66.4} &  \textbf{48.1}& \textbf{24.5}& \textbf{47.2}& \textbf{62.7}&2.5\\
    
    \Xhline{2\arrayrulewidth}
    \multicolumn{15}{@{}l}{
    M R-CNN \citep{MaskRcnn}; HTC \citep{htc}; RTMD \citep{lyu2022rtmdet}; QueryInst \citep{queryinstances};} \\
    \multicolumn{15}{@{}l}{
    R50-FPN, R101-FPN \citep{resnet,FPN}; X101-FPN \citep{resnext,FPN};} \\
    \multicolumn{15}{@{}l}{
    CX-T-FPN \citep{liu2022convnext, FPN};
    Swin \citep{liu2021swin};
    Vit \citep{li2022VitDet};} \\
    \multicolumn{15}{@{}l}{
    CX-v2-FPN \citep{woo2023convnextv2, FPN};
    CSPX-PAFPN \citep{wang2020cspnet,resnext,liu2018path}; 
    }
    
    \end{tabular}
    \end{center}
    \label{tab: main results on COCO Dataset}
\end{table*}

\begin{figure*}[!t]
\centering
\begin{minipage}[t]{0.97\textwidth}
\centering
\includegraphics[width=1\textwidth]{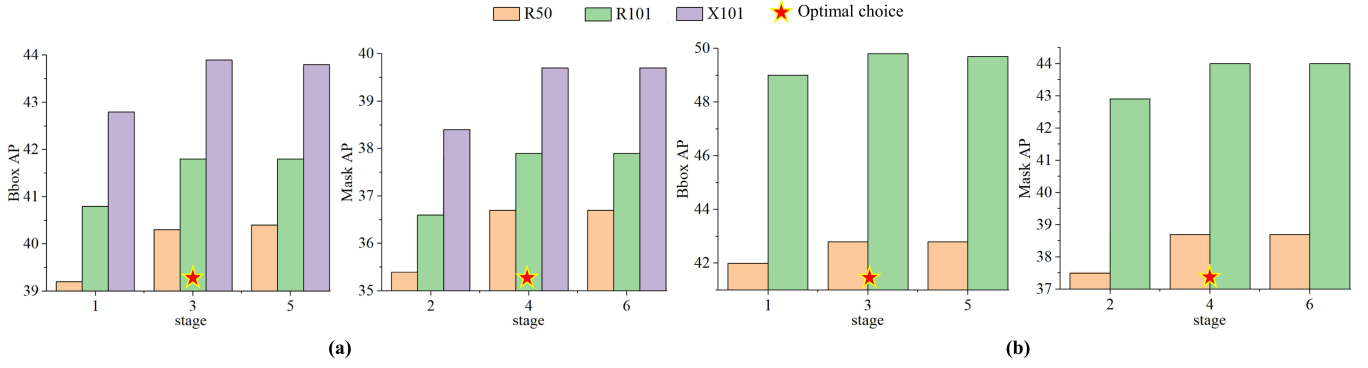}
\end{minipage}
\vspace{-2mm}
\caption{
The results at different stages in the turbo-inference process with models Mask R-CNN(a) and QueryInst(b). Left: detection results at stages 1, 3 and 5. Right: segmentation results at stages 2, 4 and 6. Star symbols denote our optimal choices. The results of these models presented in Tables \ref{tab: main results on COCO Dataset}, \ref{tab: main results on Cityscapes Dataset} and \ref{tab: main results on IFLYTEK Dataset} were obtained with these choices.
}
\label{Mask}
\end{figure*}

\begin{figure*}[!t]
\centering
\begin{minipage}[t]{0.97\textwidth}
\centering
\includegraphics[width=1\textwidth]{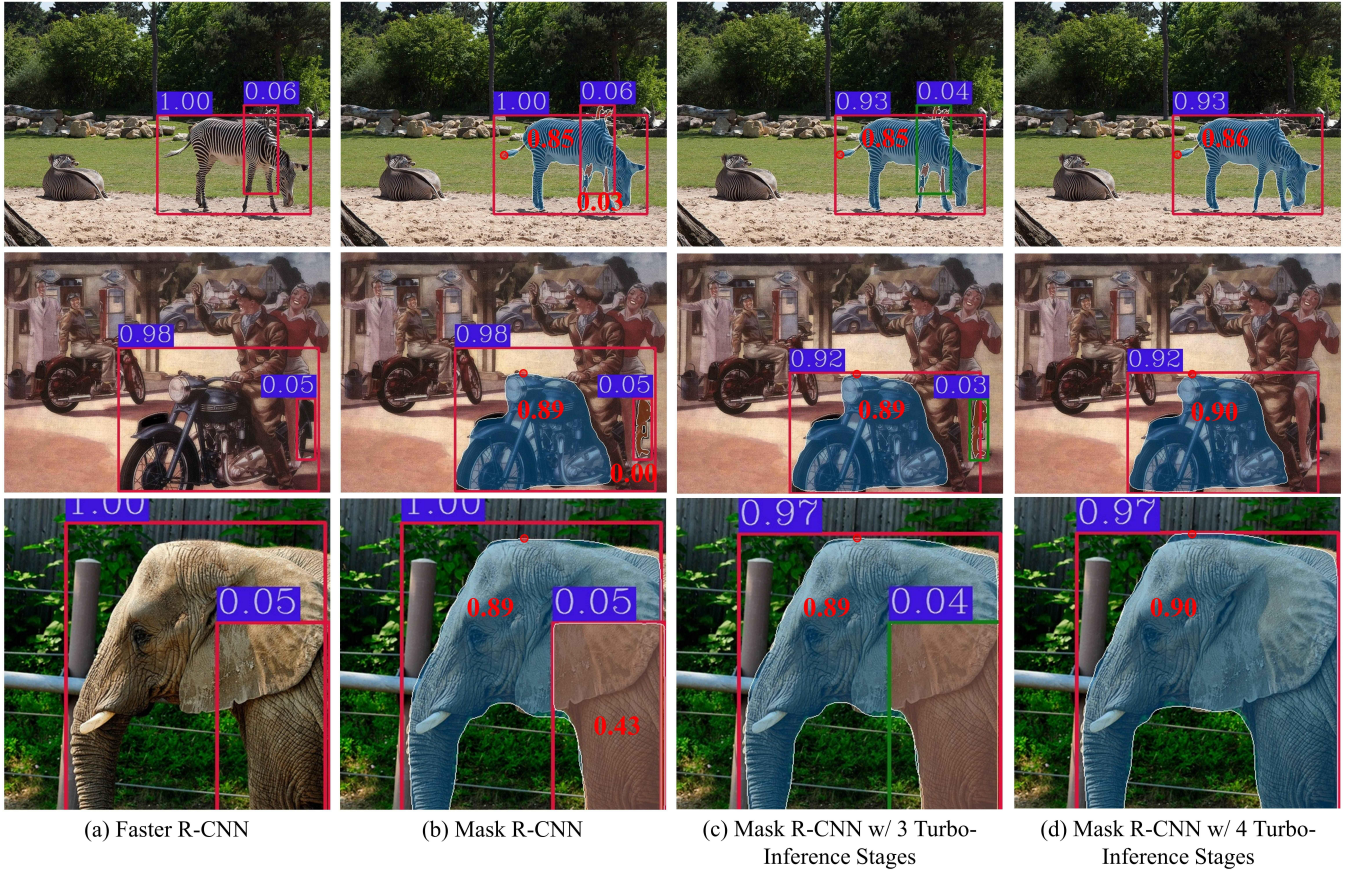}
\end{minipage}
\caption{
COCO example result tuples from (a) Vanilla Faster R-CNN, (b) Vanilla Mask R-CNN, (c) Mask R-CNN with 3 turbo-inference stages, and (d) Mask R-CNN with 4 turbo-inference stages. All symbols have the same meanings as in Figure \ref{coco into}. Green boxes indicate those removed by \emph{turbo-inference}.
}
\label{coco example}
\end{figure*}

\section{Experiments}

\subsection{Datasets and metrics.}
\noindent\textbf{Datasets.} We conducted experiments on the publicly available COCO~\citep{lin2014microsoft} and Cityscapes~\citep{cordts2016cityscapes} datasets and the private iFLYTEK~\citep{zhao2022winning} dataset. The COCO dataset comprises 118k training images and 5k validation images, covering 80 object categories with instance-level annotations. The Cityscapes dataset consists of 2975, 500, and 1525 images of size 1024×2048 for training, validation, and test sets, respectively. It contains objects of 8 categories. The iFLYTEK dataset is a remote sensing dataset for instance segmentation. It includes high-resolution remote sensing images illustrating cultivated lands of a single category. The training and validation sets consist of 16 and 15 huge images, which were converted into small overlapped images of size 512×512 following~\citep{zhao2022winning}, resulting in 3744 training and 3879 validation clips, respectively.

\noindent \textbf{Metrics.} The standard COCO-style $\rm AP$ metric~\citep{lin2014microsoft} was employed as the evaluation metric for both object detection and instance segmentation tasks in all three datasets. For inference speed, we measured the frames per second (FPS) on one RTX 2080Ti GPU with batch size 2.

\subsection{Implementation details}

We implemented our approach based on the MMDetection toolboxes~\citep{chen2019mmdetection} with default hyper-parameters and configurations, except for the newly designed parts. In the Box refinement module, we set $S_{B}$ to 0.23 to ensure that the refined RoI regions generated by the module include uncertain regions with ambiguous foreground confidence as much as possible.
In the Maskness module, we set $S_{M}$ to 0.5 for Mask R-CNN and HTC, and 0.2 for QueryInst.
Setting $S_{M}$ to 0.2 for QueryInst yielded better results than setting it to 0.5. 

\subsection{Main results}

We evaluated our turbo-inference strategy based on different methods with different backbones (Table~\ref{tab: main results on COCO Dataset}). 
These methods include the classical Mask R-CNN, query-based QueryInst, HTC utilizing a two-stage detector, and RTMDet using a one-stage detector. For Mask R-CNN and HTC, we employed a 4-stage turbo-inference, while for RTMDet and QueryInst, a 3-stage turbo-inference was utilized (Section IV. D).
In various configurations, models employing the turbo-inference strategy outperformed their counterparts in both detection and segmentation tasks, all while sacrificing only a certain amount of inference time.

Based on Mask R-CNN, the model equipped with our turbo-inference strategy substantially improved detection and segmentation accuracy, achieving improvements of 1.1\% box AP and 1.3\% mask AP over the baseline when using ResNet-50-FPN as the backbone. Introducing the turbo-inference strategy degraded $AP_{S}^m$ but boosted $AP_L^m$. 
These were achieved at the cost of a slight decrease in FPS, dropping from 15.7 to 12.0.

On HTC (utilizing a two-stage detector) and RTMDet (utilizing a one-stage detector), our turbo-inference strategy also achieved significant improvements.
Based on the HTC which incorporates a multi-stage refinement strategy, the model equipped with the turbo-inference strategy yielded 0.4\% box AP and 0.9\% mask AP gains compared with its counterpart, using ResNet-50-FPN as the backbone. 
On RTMDet-m, compared to the baseline using CSPResNeXt-PAFPN\citep{wang2020cspnet, liu2018path, resnext} as the backbone, the model equipped with the turbo-inference strategy gained 0.5\% in box AP and 0.3\% in mask AP.
The improvements achieved by turbo-inference became slightly smaller due to the powerful localization and segmentation capabilities of these methods.
Additionally, both of them experienced a 1.0 FPS point drop.

Based on the query-based method QueryInst, the models with turbo-inference still achieved considerable gains. Using ResNet-50-FPN as a backbone, the model equipped with turbo-inference achieved improvements of 0.8\% box AP and 1.2\% mask AP, while experiencing a slight drop of 1.5 FPS. The AP gains achieved on QueryInst exceeded the gains on HTC, which may be because QueryInst is designed to predict positive results directly without any post-processing schemes, \textit{e.g.}, NMS. Although NMS is unsuitable for query-based methods, our experiments showed that a more efficient post-processing scheme can still achieve substantial gains.

When scaled up to more powerful backbones ResNet-101, ResNeXt-101~\citep{resnext}, Swin-transformer\citep{liu2021swin}, Vision-transformer\citep{li2022VitDet}, ConvNeXt\citep{liu2022convnext} and ConvNeXt-v2\citep{woo2023convnextv2}, the models with the turbo-inference strategy still yielded significant gains, demonstrating the generality of our method.

\begin{table}[htbp]
\caption{ Effect of the turbo-inference on Mask R-CNN with ResNet-50-FPN Backbone.}
\begin{center}
\begin{tabular}{c|c|c|c|c}
\Xhline{2\arrayrulewidth}
Box refinement & Maskness & Turbo-seg & $\rm AP^b$ & $\rm AP^m$ \\
\arrayrulecolor{black}\hline
           &            &             &39.9 &35.6 \\
\checkmark &            &             &40.7 &35.6 \\
\checkmark &            & \checkmark  &40.7 &35.8 \\
\checkmark & \checkmark & \checkmark  &41.0 &36.9 \\
\Xhline{2\arrayrulewidth}
\end{tabular}
\end{center}
\label{tab: ablation study of TaskCascade}
\end{table}

\begin{table}[htbp]
\caption{ Effect of the turbo-inference on QueryInst with ResNet-50-FPN Backbone.}
\begin{center}
\begin{tabular}{c|c|c|c}
\Xhline{2\arrayrulewidth}
Box refinement & Maskness & $\rm AP^b$ & $\rm AP^m$ \\
\arrayrulecolor{black}\hline
 & &42.0 &37.5 \\
\checkmark &  &42.2 &37.5 \\
\checkmark &\checkmark  &42.8 &38.7 \\
\Xhline{2\arrayrulewidth}
\end{tabular}
\end{center}
\label{tab: ablation study of FastTaskCascade}
\end{table}

\begin{table*}[htbp]
\caption{Results on the Cityscapes validation dataset. Turbo is short for turbo-inference.}

\begin{center}
\begin{tabular}{c |c|ccccc|ccccc}

\Xhline{2\arrayrulewidth}
Method & Backbone & $\rm AP^b$ & $\rm AP_{50}^b$ & $\rm AP_{S}^b$ & $\rm AP_{M}^b$ & $\rm AP_{L}^b$ & $\rm AP^m$ & $\rm AP_{50}^m$ & $\rm AP_{S}^m$ & $\rm AP_{M}^m$ & $\rm AP_{L}^m$\\

\arrayrulecolor{black}\hline

Mask R-CNN & R50-FPN &40.9 & \textbf{65.7} &17.5 &40.0 &62.4 &36.7&62.1 &\textbf{10.8}&32.2&60.1 \\

w/ Turbo& R50-FPN& \textbf{41.9} & 64.5 &\textbf{17.6} &  \textbf{40.6}&  \textbf{63.7} & \textbf{38.1} &\textbf{62.1} &10.7&\textbf{33.3} &\textbf{62.4}\\

\arrayrulecolor{black}\hline

Mask R-CNN & R101-FPN & 41.8 & 67.8 & 17.2 & 42.6 & 63.8 & 36.5&63.2&9.0&32.7&61.0 \\

w/ Turbo       & R101-FPN& \textbf{42.7} & \textbf{68.1} &\textbf{17.5 }&  \textbf{43.1} &  \textbf{65.4}& \textbf{38.1} &\textbf{63.5}&\textbf{9.4}&\textbf{33.4}&\textbf{63.9}\\

\Xhline{2\arrayrulewidth}
\end{tabular}
\end{center}
\label{tab: main results on Cityscapes Dataset}
\end{table*}

\begin{table*}[htbp]
\caption{Results on the IFLYTEK validation dataset. Turbo is short for turbo-inference.}

\begin{center}
\begin{tabular}{c |c|cccccc|cccccc}

\Xhline{2\arrayrulewidth}

Method & Backbone & $\rm AP^b$ & $\rm AP_{50}^b$ & $\rm AP_{75}^b$ & $\rm AP_{S}^b$ & $\rm AP_{M}^b$ & $\rm AP_{L}^b$ & $\rm AP^m$ & $\rm AP_{50}^m$ & $\rm AP_{75}^m$ & $\rm AP_{S}^m$ & $\rm AP_{M}^m$ & $\rm AP_{L}^m$\\




\arrayrulecolor{black}\hline


Mask R-CNN& R50-FPN &37.6&\textbf{61.9}&39.1&16.7&42.5&45.5&34.1&57.6&35.5&13.8&37.8&41.5 \\

w/ Turbo& R50-FPN      &\textbf{38.6} &61.8 &\textbf{40.1}& \textbf{17.6} &\textbf{43.3} &\textbf{46.8} &\textbf{35.9} &\textbf{58.2} &\textbf{37.4}&\textbf{14.7}&\textbf{39.2}&\textbf{43.9} \\

\arrayrulecolor{black}\hline

Mask R-CNN& R101-FPN&37.3 &61.0 &39.0 &16.3 &42.1 &45.8 &34.2 &57.2 &35.3 &13.6 &37.7 &42.3 \\

w/ Turbo& R101-FPN      &\textbf{38.6} &\textbf{61.1} &\textbf{40.1} &\textbf{17.0} &\textbf{43.1} &\textbf{47.5} &\textbf{35.9} &\textbf{57.9} &\textbf{37.2} &\textbf{14.4} &\textbf{39.2} &\textbf{44.6} \\

\arrayrulecolor{black}\hline

HTC& R50-FPN&40.9 &\textbf{63.5} &43.3 &18.4 &45.5 &49.4 &37.6 &60.5 &39.8 &14.9 &41.3 &46.0 \\

w/ Turbo& R50-FPN&\textbf{41.4} &63.4 &\textbf{43.6} &\textbf{18.9} &\textbf{45.8} &\textbf{49.9} &\textbf{38.6} &\textbf{60.8} &\textbf{40.8} &\textbf{15.8} &\textbf{41.9} &\textbf{47.3} \\

\arrayrulecolor{black}\hline

QueryInst& R50-FPN&28.1 &46.9 &28.9 &11.3 &31.6 &34.1 &26.1 &44.0 &26.9 &9.6 &29.2 &33.0 \\

w/ Turbo& R50-FPN&\textbf{29.7} &\textbf{47.7} &\textbf{30.8} &\textbf{12.5} &\textbf{33.0} &\textbf{36.1} &\textbf{27.6} &\textbf{45.1} &\textbf{28.7} &\textbf{10.4} &\textbf{30.2} &\textbf{35.0}\\

\Xhline{2\arrayrulewidth}
\end{tabular}
\end{center}
\label{tab: main results on IFLYTEK Dataset}
\end{table*}

\begin{table*}[!t]
    \caption{Combination of Soft NMS and our turbo-inference strategy on the COCO validation dataset.}

    \begin{center}
        \begin{tabular}{c |c|cccccc|cccccc|c}
    
    \Xhline{2\arrayrulewidth}
    
    Method & Backbone & $\rm AP^b$ & $\rm AP_{50}^b$ & $\rm AP_{75}^b$ & $\rm AP_{S}^b$ & $\rm AP_{M}^b$ & $\rm AP_{L}^b$ & $\rm AP^m$ & $\rm AP_{50}^m$ & $\rm AP_{75}^m$ & $\rm AP_{S}^m$ & $\rm AP_{M}^m$ & $\rm AP_{L}^m$ & FPS\\
    
    
    
    
    
    \arrayrulecolor{black}\hline
    
    M R-CNN & & 40.8 &61.0&  44.5 &23.0 &45.0 &54.1&36.6&  57.9&  39.1 &\textbf{19.2} & 40.2 & 50.5&\textbf{13.5} \\
    w/ S NMS& R101-FPN&41.6&\textbf{61.1}&45.8&23.5&45.8&54.9&36.9&58.0&39.6&17.0&40.0&54.6&13.3\\
    
    w/ Turbo& & 41.8 & 60.7&  45.5& 23.4& 45.9&  55.9& 37.9&  58.0&   40.8& 17.5& 41.0& 55.3& 9.8\\
    
    w/ Both& &\textbf{42.5}&60.7&\textbf{46.6}&\textbf{23.9}&\textbf{46.7}&\textbf{56.8}&\textbf{38.2}&\textbf{58.1}&\textbf{41.2}&17.6&\textbf{41.2}&\textbf{55.8}& 9.0\\
    
    \arrayrulecolor{black}\hline
    
    
    
    
    
    
    
    
    
    
    
    HTC & & 44.8& \textbf{63.3} & 48.8& 25.7& 48.5& 60.2&39.6&  61.0 & 42.8&\textbf{21.3}& 42.9& 55.0&\textbf{5.2} \\
    
    w/ S NMS& R101-FPN& 45.3& 63.2&49.9&\textbf{26.0}&49.1&60.8&39.8&60.8&43.2&18.5&42.9&58.8&5.1\\
    
    w/ Turbo& & 45.1 & 63.3 &  48.9 &25.6 &48.8 & 60.8 &40.5 &  \textbf{61.1} & 43.7 & 18.9 & 43.5 &59.4 &4.3\\
    
    w/ Both& & \textbf{45.6} &63.1 &\textbf{50.0} &25.9 &\textbf{49.4}&\textbf{61.4} &\textbf{40.6} &61.0 &\textbf{44.1} &18.9 &\textbf{43.6}&\textbf{59.7}&4.2 \\

    \arrayrulecolor{black}\hline

    RTMDet-m &  & 48.8  & \textbf{66.8}  & 53.3& 29.9 & 54.0 & 65.1 &42.1 &  63.8  & 45.1 &19.3 & 46.4 & 63.1&\textbf{3.7} \\

    w/ S NMS & CSPX- & 48.9  & 66.8  & 53.4& \textbf{30.0} & 54.1 & 65.2 &42.1 &  \textbf{63.9}  & 45.2 &19.4 & 46.4 & 63.2 &3.2 \\

    w/ Turbo & PAFPN & 49.3  & 66.7 &  53.5 &29.9 & 54.5 & 66.1 &42.4 &  63.8 &  45.5& 19.5& 46.7& 63.3&2.7\\

    w/ Both & & \textbf{49.3}  & 66.7 &  \textbf{53.6} &29.9 & \textbf{54.5} & \textbf{66.2} &\textbf{42.4} &  63.8 &  \textbf{45.5}& \textbf{19.5}& \textbf{46.7}& \textbf{63.4}&2.4\\

    \arrayrulecolor{black}\hline
    
    \Xhline{2\arrayrulewidth}
    \end{tabular}
    \end{center}
    \label{tab: Soft NMS and Turbo results on COCO Dataset}
\end{table*}

\subsection{Quantitative results}

We present some visualization examples from COCO in Figure \ref{coco example}. In examples shown in Figure \ref{coco example}(b), detection boxes predicted by the vanilla Mask R-CNN contain a significant amount of redundant space, with redundant boxes on the same objects.
In contrast, Mask R-CNN with turbo-inference generates more precise bounding boxes and refines classification scores to remove redundant boxes with scores below 0.05 in the third stage (Figure \ref{coco example}(c)); then, it predicts higher-quality masks in the fourth stage (Figure \ref{coco example}(d)).

\subsection{Ablation experiments}

\subsubsection{Different numbers of stages for turbo-inference} We first comprehensively evaluated the performance of the turbo-inference strategy with different numbers of stages on the COCO dataset. 
In the models mentioned above, Mask R-CNN and HTC employ vanilla convolutional layers for segmentation, while RTMDet and QueryInst utilize dynamic convolutional layers. Dynamic convolutional layers, incorporating additional proposal features alongside RoI features extracted from detection boxes, may interfere with turbo-inference. We chose Mask R-CNN and QueryInst to represent these two cases for detailed ablation experiments.

The results are shown in Figure \ref{Mask}. Stages 1, 3, and 5 were dedicated to detection tasks, while stages 2, 4, and 6 were dedicated to segmentation tasks. Classification scores were refined only in stage 3. Subsequent detection stages only involved refining detection boxes.

For Mask R-CNN, in stage 3, the detector achieved 1.1\% box AP improvement compared with the baseline, using ResNet-50-FPN as the backbone, while the refinement of classification scores further enhanced the mask AP by 1.2\%.  Stage 4, which involved the segmentation head using more accurate refined detection boxes, improved the mask AP by 0.1\%. However, stages 5 and 6 did not bring significant AP gains despite the refinement of detection boxes and prediction masks. Therefore, we adopted four stages in the turbo-inference pipeline for Mask R-CNN and HTC by default.

For QueryInst, using ResNet-101-FPN as a backbone, in stage 3, the detector achieved 0.8\% improvement in box AP.  The segmentation network also benefited from the refinement of classification scores in this stage, leading to 1.2\% mask AP improvement. However, in stage 4, there was a slight decrease of 0.1\% in mask AP due to disruptions in the interaction process driven by the attention mechanism. Consequently, we adopted three stages in the turbo-inference pipeline for QueryInst by default.

In contrast to Mask R-CNN, QueryInst suffered from a decrease in Mask AP during stage 4 of turbo-inference. The decrease could be attributed to interference from the turbo-inference framework with the dynamic convolution interaction process in QueryInst. QueryInst generates high-dimensional proposal features as object queries, each corresponding one-to-one with bounding box/mask RoI features, playing a crucial role in outputting final object features. However, turbo-inference can only refine the mask RoI features and not the proposal features. This asymmetry disrupts the interaction process when the refined mask RoI features are introduced, resulting in a loss of segmentation accuracy.

\subsubsection{Different modules in turbo-inference} We also evaluated the influence of various modules within turbo-inference on performance. Tables~\ref{tab: ablation study of TaskCascade} and \ref{tab: ablation study of FastTaskCascade} presented ablation experiments conducted on COCO $val2017$ using the Mask R-CNN and QueryInst, respectively, with ResNet-50-FPN as the backbone.

From Table~\ref{tab: ablation study of TaskCascade}, we first observe that the Box refinement module generated more accurate detection boxes by using predicted masks, which improved the box AP by 0.8\%. Second, the refined bounding boxes improved object localization accuracy, consequently assisting the turbo-segmentation head in generating more precise predicted masks, resulting in a 0.2\% increase in mask AP. Third, the Maskness module refined classification scores by incorporating the uncertainty of predicted masks so that classification scores reflected both localization and segmentation confidence, resulting in 0.3\% improvement in box AP and 1.1\% in mask AP. 

From Table~\ref{tab: ablation study of FastTaskCascade}, we first observe that the Box refinement module only improved box AP by 0.2\% because QueryInst employed a strong Sparse R-CNN detector. Second, The Maskness module yielded gains of 0.6\% in box AP and 1.2\% in mask AP. Compared with HTC, the Maskness module achieved larger gains in box AP based on QueryInst, which was because QueryInst did not use NMS to adjust classification scores, making them less correlated with localization confidence.

As shown in Tables~\ref{tab: ablation study of TaskCascade} and~\ref{tab: ablation study of FastTaskCascade}, the Box refinement module mainly enhanced object detection, but the gains in box AP became smaller for methods with a multi-stage refinement strategy. After the Box refinement module came into play, the turbo-segmentation head could slightly boost mask AP. The Maskness module, which mainly boosted mask AP, also significantly improved box AP when applied to the query-based QueryInst. 

\subsection{Experiments on Cityscapes}

We validated our turbo-inference strategy on the Cityscapes dataset (Table~\ref{tab: main results on Cityscapes Dataset}). Following the default settings in MMDetection, the dataset was replicated eight times. All models were initialized from the corresponding COCO pre-trained models and then fine-tuned on the fine annotations for 8 epochs with multi-scale data augmentations. The hyperparameters in the turbo-inference strategy were kept the same as on COCO. 

We conducted experiments based on the Mask R-CNN using ResNet-50-FPN and ResNet-101-FPN as backbones. The models with turbo-inference strategy achieved consistent improvements on different baselines, further demonstrating the effectiveness of our approach.

\subsection{Experiments on iFLYTEK}

We conducted experiments on the remote sensing dataset iFLYTEK. All images were padded to (600,600). All other settings were retained the same as on COCO. 

Table~\ref{tab: main results on IFLYTEK Dataset} shows that the models equipped with our turbo-inference strategy yielded consistent gains on the iFLYTEK dataset as on the COCO dataset. Specifically, the model with turbo-inference strategy achieved significant improvements on Mask R-CNN. With ResNet-50-FPN as the backbone, the models with turbo-inference achieved improvements of 1.0\% in box AP and 1.8\% in mask AP over the Mask R-CNN counterpart.

\subsection{Combination with Soft NMS}

Soft NMS~\citep{bodla2017soft} is a popular training-free method to improve object detection and instance segmentation by adjusting classification scores. Soft NMS attenuates classification scores based on IoUs among bounding boxes to address the inconsistency between classification scores and localization accuracy. This results in improvements in box AP but has little effect on mask AP. We have observed that our turbo-inference strategy was compatible with Soft NMS, and the improvements achieved by both methods could be accumulated, leading to significant gains in both box and mask AP.

Table \ref{tab: Soft NMS and Turbo results on COCO Dataset} demonstrates that when combined with Soft NMS, our turbo-inference strategy still achieved similar gains and yielded superior results on Mask R-CNN, HTC, and RTMDet.

\section{Conclusion}

We present a training-free pipeline called turbo-inference to improve both object detection and instance segmentation. Unlike prior approaches, which usually consider instance segmentation as a downstream task of object detection, our turbo-inference strategy introduces the additional turbo-detection head and the turbo-segmentation head to establish a closed loop between object detection and instance segmentation. It enables the iterative utilization of complementary information between both tasks. Our turbo-inference strategy can be easily integrated into existing instance segmentation methods that follow the \emph{detect-then-segment} pipeline, such as the Mask R-CNN, HTC, and QueryInst. Extensive experiments on the COCO, Cityscapes, and iFLYTEK datasets demonstrate the effectiveness of our method. 

The main drawback of our method is its compatibility exclusively with top-down frameworks. Another limitation is a reduction in inference speed. However, this decrease is not significant when employing a large backbone network since the detection and segmentation heads are typically much smaller than the backbone network.

\section*{Acknowledgments}
This work was supported by the Opening Foundation of Key Laboratory of Advanced Manufacture Technology for Automobile Parts, Ministry of Education (No. 2021KLMT05) and the National Natural Science Foundation of China (No. 51975057).\\

\bibliographystyle{model2-names}
\bibliography{refs}

\end{document}